\documentclass[10pt,twocolumn,letterpaper]{article}
\usepackage{cvpr}
\usepackage{times}
\usepackage{epsfig}
\usepackage{graphicx}
\usepackage{amsmath}
\usepackage{amssymb}
\usepackage{booktabs}
\usepackage{tablefootnote}

\usepackage{verbatim}
\usepackage{mathtools}
\usepackage{makecell}
\usepackage{multirow}
\usepackage{color,soul}

\DeclarePairedDelimiter{\ceil}{\lceil}{\rceil}

\newcommand*\samethanks[1][\value{footnote}]{\footnotemark[#1]}

\usepackage[pagebackref=true,breaklinks=true,letterpaper=true,colorlinks,bookmarks=false]{hyperref}

\cvprfinalcopy 

\def\cvprPaperID{37} 

\ifcvprfinal\fi
\begin{document}

\title{End-to-End Dense Video Captioning with Masked Transformer}

\author{Luowei Zhou\thanks{Equal contribution}\\
University of Michigan\\
{\tt\small luozhou@umich.edu}
\and
Yingbo Zhou\samethanks{}\\
Salesforce Research\\
{\tt\small yingbo.zhou@salesforce.com}
\and
Jason J. Corso\\
University of Michigan\\
{\tt\small jjcorso@eecs.umich.edu}
\and
Richard Socher\\
Salesforce Research\\
{\tt\small richard@socher.org}
\and
Caiming Xiong\thanks{Corresponding author}\\
Salesforce Research\\
{\tt\small cxiong@salesforce.com}
}

\maketitle

\begin{abstract}
Dense video captioning aims to generate text descriptions for all events in an untrimmed video. This involves both detecting and describing events. Therefore, all previous methods on dense video captioning tackle this problem by building two models, \ie an event proposal and a captioning model, for these two sub-problems. The models are either trained separately or in alternation. This prevents direct influence of the language description to the event proposal, which is important for generating accurate descriptions. To address this problem, we propose an end-to-end transformer model for dense video captioning. The encoder encodes the video into appropriate representations. The proposal decoder decodes from the encoding with different anchors to form video event proposals. The captioning decoder employs a masking network to restrict its attention to the proposal event over the encoding feature. This masking network converts the event proposal to a differentiable mask, which ensures the consistency between the proposal and captioning during training. In addition, our model employs a \emph{self-attention} mechanism, which enables the use of efficient non-recurrent structure during encoding and leads to performance improvements. We demonstrate the effectiveness of this end-to-end model on ActivityNet Captions and YouCookII datasets, where we achieved 10.12 and 6.58 METEOR score, respectively.
\end{abstract}

\section{Introduction}
\label{sec:intro}

Video has become an important source for humans to learn and acquire
knowledge (\eg video lectures, making sandwiches~\cite{kuehne2014language}, changing tires~\cite{alayrac2016learning}).
Video content consumes high cognitive bandwidth, and thus is slow for humans to digest. 
Although the visual signal itself can sometimes disambiguate certain semantics, one way to make video content more easily and rapidly understood by humans is to compress it in a way that retains the semantics.  
This is particularly important given the massive amount of video being produced everyday. 
Video summarization \cite{zhang2016video} is one way of doing this, but it loses the language components of the video, which are particularly important in instructional videos.
Dense video captioning~\cite{krishna2017dense}---describing  events in the video with descriptive natural language---is another way of achieving this compression while retaining the language components.

\begin{figure}[t]
\centering 
   \includegraphics[width=1\columnwidth]{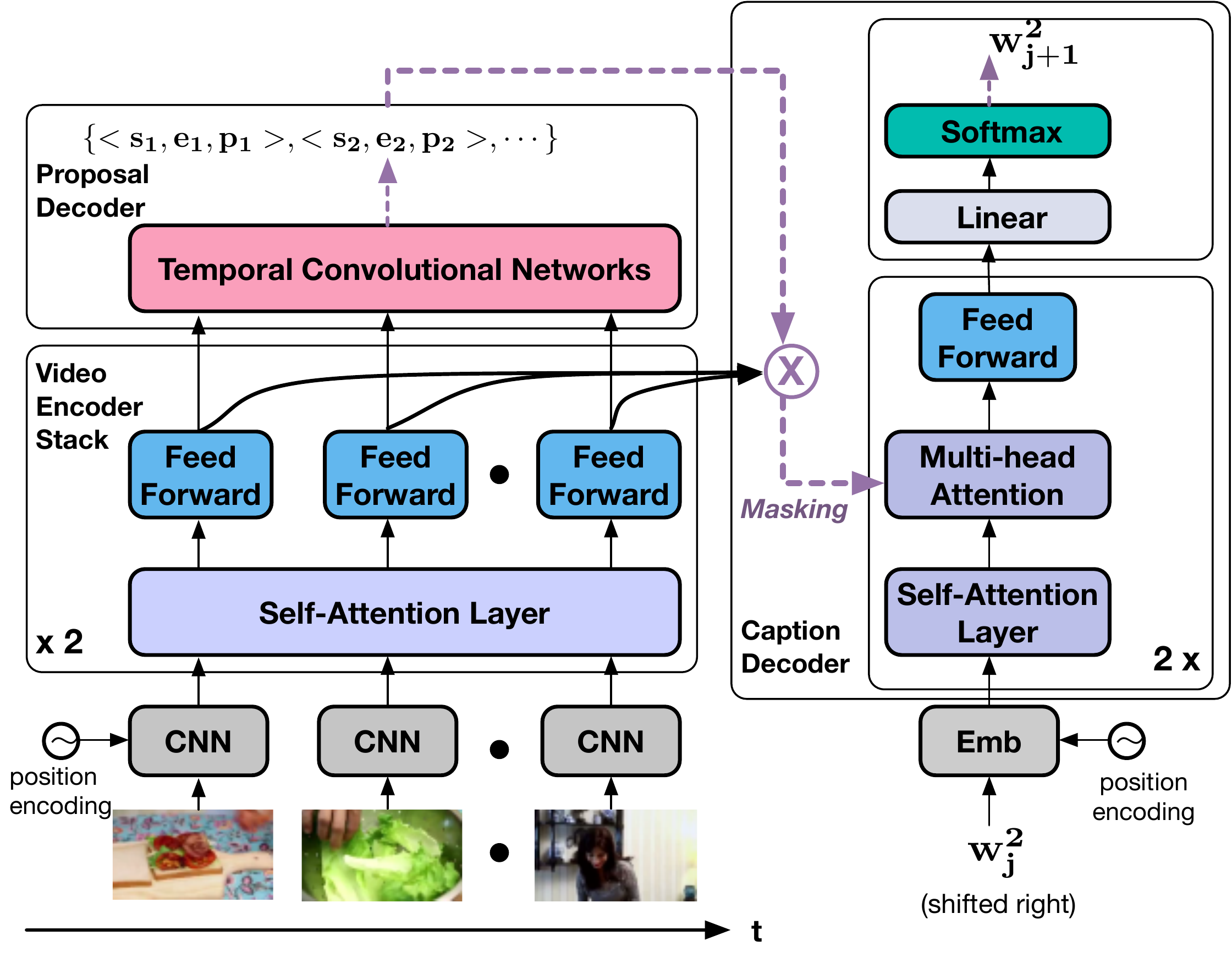}
    \vspace{-10pt}
       \caption{Dense video captioning is to localize (temporal) events from a video, which are then described with natural language sentences. We leverage temporal convolutional networks and self-attention mechanisms for precise event proposal generation and captioning.}
    \vspace{-10pt}
\label{fig:fig1}
\end{figure}

Dense video captioning can be decomposed into two parts: event detection and event description. Existing methods tackle these two sub-problems using event proposal and captioning modules, and exploit two ways to combine them for dense video captioning. 
One way is to train the two modules independently and generate descriptions for the best event proposals with the best captioning model~\cite{ghanem2017activitynet}. 
The other way is to alternate training~\cite{krishna2017dense} between the two modules, \ie, alternate between i) training the proposal module only and ii) training the captioning module on the positive event proposals while fine-tuning the proposal module.
However, in either case, the language information cannot have direct impacts on the event proposal.

Intuitively, the video event segments and language are closely related and the language information should be able to help localize events in the video. 
To this end, we propose an encoder-decoder based end-to-end model for doing dense video captioning (see Fig. \ref{fig:fig1}). The encoder encodes the video frames (features) into the proper representation. The proposal decoder then decodes this representation with different anchors to form event proposals, \ie, start and end time of the event, and a confidence score. The captioning decoder then decodes the proposal specific representation using a masking network, which converts the event proposal into a differentiable mask. This continuous mask enables both the proposal and captioning decoder to be trained consistently, \ie the proposal module now learns to adjust its prediction based on the quality of the generated caption. In other words, the language information from caption now is able to guide the visual model to generate more plausible proposals.
In contrast to the existing methods where the proposal module solves a class-agnostic binary classification problem regardless the details in the video content, our model enforces the consistency between the content in the proposed video segment and the semantic information in the language description.

Another challenge for dense video captioning, and more broadly for sequence modeling tasks, is the need to learn a representation that is capable of capturing long term dependencies. Recurrent Neural Networks (RNN) are possible solutions to this problem, however, learning such representation is still difficult \cite{pascanu2013difficulty}. \emph{Self-attention} \cite{lin2017structured,paulus2017deep,vaswani2017attention} allows for an attention mechanism within a module and is a potential way to learn this long-range dependence. In self-attention the higher layer in the same module is able to attend to all states below it. This made the length of the paths of states from the higher layer to all states in the lower layer to be one, and thus facilitates more effective learning. The shorter path length facilitates learning these dependencies because larger gradients can now pass to all states. Transformer~\cite{vaswani2017attention} implements a fast self-attention mechanism and has demonstrated its effectiveness in machine translation. Unlike traditional sequential models, transformer does not require unrolling across time, and therefore trains and tests much faster as compared to RNN based models. We employ transformer in both the encoder and decoder of our model.

Our main contributions are twofold. First, we propose an end-to-end model for doing dense video captioning. A differentiable masking scheme is proposed to ensure the consistency between proposal and captioning module during training. Second, we employ self-attention: a scheme that facilitates the learning of long-range dependencies to do dense video captioning. To the best of our knowledge, our model is the first one that does not use a RNN-based model for doing dense video captioning. In addition, we achieve competitive results on ActivityNet Captions~\cite{krishna2017dense} and YouCookII~\cite{zhou2017procnets} datasets.

\section{Related Work}
\label{sec:related}

\noindent\textbf{Image and Video Captioning.}\quad In contrast to earlier video captioning papers, which are based on models like hidden Markov models and ontologies \cite{yu2013grounded, das2013thousand}, recent work on captioning is dominated by deep neural network-based methods~\cite{vinyals2015show,xu2015show,you2016image,zhou2016watch,yao2016boosting,rennie2016self}. 
Generally, they use Convolutional Neural Networks (CNNs) \cite{simonyan2014very,he2016deep} for encoding video frames, followed by a recurrent language decoder, e.g., Long Short-Term Memory~\cite{hochreiter1997long}.
They vary mainly based on frame encoding, e.g., via mean-pooling~\cite{venugopalan2014translating,gan2016semantic}, recurrent nets~\cite{donahue2015long,venugopalan2015sequence}, and attention mechanisms~\cite{yao2015describing,pan2016video,gan2016semantic}.
The attention mechanism was initially proposed for machine translation~\cite{bahdanau2014neural} and has achieved top performance in various language generation tasks, either as temporal attention~\cite{yao2015describing}, semantic attention~\cite{gan2016semantic} or both~\cite{pan2016video}. 
Our work falls into the first of the three types. In addition to using cross-module attention, we apply self-attention~\cite{vaswani2017attention} within each module. 

\noindent\textbf{Temporal Action Proposals.}\quad Temporal action proposals (TAP) aim to temporally localize action-agnostic proposals in a long untrimmed video. Existing methods formulate TAP as a binary classification problem and differ in how the proposals are proposed and discriminated from the background. 
Shuo et al.~\cite{shou2016temporal} 
propose and classify proposal candidates directly over video frames in a sliding window fashion, which is computationally expensive. More recently, inspired by the anchoring mechanism from object detection~\cite{ren2015faster}, two types of methods have been proposed---explicit anchoring~\cite{gao2017turn,zhou2017procnets} and implicit anchoring~\cite{escorcia2016daps,buch2017sst}. 
In the former case, each anchor is an encoding of the visual features between the anchor temporal boundaries and is classfied as action or background. In implicit anchoring, recurrent networks encode the video sequence and, at each anchor center, multiple anchors with various sizes are proposed based on the same visual feature. 
So far, explicit anchoring methods accompanied with location regression yield better performance~\cite{gao2017turn}. Our proposal module is based upon Zhou et al.~\cite{zhou2017procnets}, which is designed to detect long complicated events rather than actions. 
We further improve the framework with a temporal convolutional proposal network and self-attention based context encoding.

\noindent\textbf{Dense Video Captioning.}\quad The video paragraph captioning method proposed by Yu \etal \cite{yu2015video} generates sentence descriptions for temporally localized video events. However, the temporal locations of each event are provided beforehand. Das et al.~\cite{das2013thousand} generates dense captions over the entire video using sparse object stitching, but their work relies on a top-down ontology for the actual description and is not data-driven like the recent captioning methods. The most similar work to ours is Krishna et al.~\cite{krishna2017dense} who introduce a dense video captioning model that learns to propose the event locations and caption each event with a sentence.
%
However, they combine the proposal and the captioning modules through co-training and are not able to take advantage of language to benefit the event proposal~\cite{heilbron2017scc}. To this end, we propose an end-to-end framework for doing dense video captioning that is able to produce proposal and description simultaneously. Also, our work directly incorporates the semantics from captions to the proposal module. 
%
\section{Preliminary}
\label{sec:preliminary}
In this section we introduce some background on Transformer~\cite{vaswani2017attention}, which is the building block for our model. 
We start by introducing the \emph{scaled dot-product attention}, which is the foundation of transformer. Given a query $q_i \in \mathbb{R}^d$ from all $T'$ queries, a set of keys $k_t \in \mathbb{R}^d$ and values $v_t \in \mathbb{R}^d$ where $t=1,2,...,T$, the scaled dot-product attention outputs a weighted sum of values $v_t$, where the weights are determined by the dot-products of query $q$ and keys $k_t$. In practice, we pack $k_t$ and $v_t$ into matricies $K=(k_1,...,k_T)$ and $V=(v_1,...,v_T)$, respectively. The attention output on query $q$ is:
\begin{equation}
A(q_i, K, V)=V\frac{\exp \left\{K^Tq_i/\sqrt{d}\right\}}{\sum_{t=1}^{T}{\exp \{k_t^Tq_i/\sqrt{d}}\} }
\end{equation}
The \emph{multi-head attention} consists of $H$ paralleled scaled dot-product attention layers called ``head'', where each ``head'' is an independent dot-product attention. The attention output from multi-head attention is as below:
\begin{align}
\text{MA}(q_i, K, V)=W^O\begin{pmatrix}
\text{head}_1 \\
\cdots \\
\text{head}_H
\end{pmatrix}\\
\text{head}_j = A(W_j^qq_i, W_j^KK, W_j^VV)
\end{align}
where $W_j^q, W_j^K, W_j^V \in\mathbb{R}^{\frac{d}{H} \times d}$ are the independent head projection matrices, $j=1,2,...,H$, and $W^O \in \mathbb{R}^{d\times d}$.

\begin{figure}[t]
\centering 
  \includegraphics[width=0.7\columnwidth]{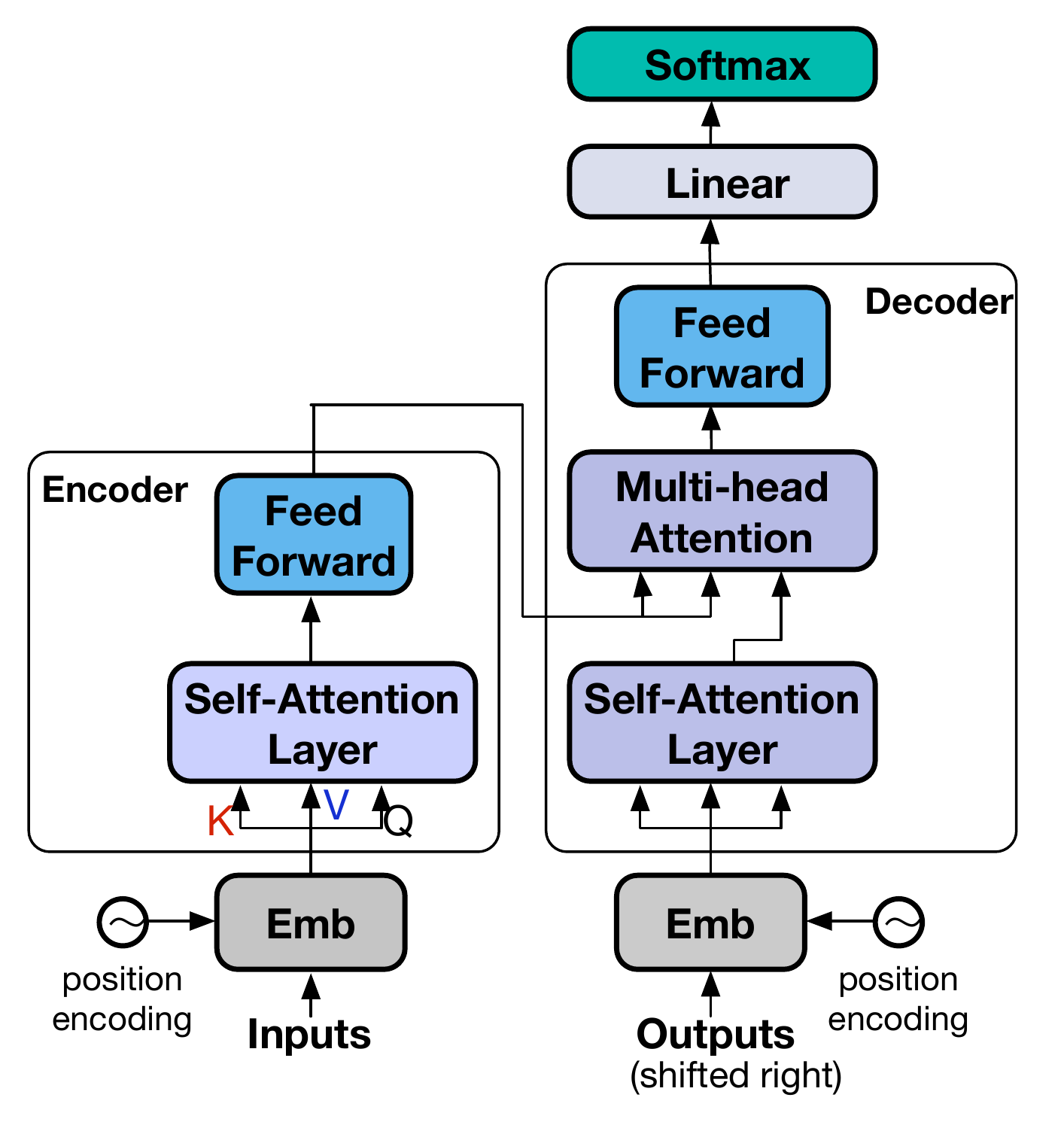}
    \vspace{-10pt}
       \caption{Transformer with 1-layer encoder and 1-layer decoder.}
    \vspace{-10pt}
\label{fig:transformer}
\end{figure}

This formulation of attention is quite general, for example when the query is the hidden states from the decoder, and both the keys and values are all the encoder hidden states, it represents the common cross-module attention.
\emph{Self-attention}~\cite{vaswani2017attention} is another case of multi-head attention where the queries, keys and values are all from the same hidden layer (see also in Fig.~\ref{fig:transformer}). 

Now we are ready to introduce Transformer model, which is an encoder-decoder based model that is originally proposed for machine translation~\cite{vaswani2017attention}. The building block for Transformer is multi-head attention and a pointwise feed-forward layer. The pointwise feed-forward layer takes the input from multi-head attention layer, and further transforms it through two linear projections with ReLU activation. The feed-forward layer can also be viewed as two convolution layers with kernel size one. The encoder and decoder of Transformer is composed by multiple such building blocks, and they have the same number of layers. The decoder from each layer takes input from the encoder of the same layer as well as the lower layer decoder output. Self-attention is applied to both encoder and decoder. Cross-module attention between encoder and decoder is also applied.
Note that the self-attention layer in the decoder can only attend to the current and previous positions to preserve the auto-regressive property. 
Residual connection~\cite{he2016deep} is applied to all input and output layers. Additionally, layer normalization~\cite{ba2016layer} (LayerNorm) is applied to all layers. 
Fig.~\ref{fig:transformer} shows a one layered transformer.

\section{End-to-End Dense Video Captioning}
\label{sec:end-to-end}
Our end-to-end model is composed of three parts: a video encoder, a proposal decoder, and a captioning decoder that contains a mask prediction network to generate text description from a given proposal. The video encoder is composed of multiple self-attention layers. The proposal decoder takes the visual features from the encoder and outputs event proposals. The mask prediction network takes the proposal output and generates a differentiable mask for a certain event proposal. To make the decoder caption the current proposal, we then apply this mask by element-wise multiplication between it, the input visual embedding and all outputs from proposal encoder. In the following sections, we illustrate each component of our model in detail.

\subsection{Video Encoder} 
\label{sec:encoder}
Each frame $x_t$ of the video $X = \{x_1,\ldots,x_T\}$ is first encoded to a continuous representation $F^0 = \{f^0_1,\ldots,f^0_T\}$. It is then fed forward to $L$ encoding layers, where each layer learns a representation $F^{l+1} = V(F^l)$ by taking input from previous layer $l$,
\begin{align}
&\text{V}(F^l) = \Psi(\text{PF}(\Gamma(F^l)),\Gamma(F^l))
\end{align}
\begin{align}
\label{eq:self_attn}
&\Gamma(F^l) = 
\begin{pmatrix}
\Psi(\text{MA}(f^l_1, F^l, F^l), f^l_1)^{\top}\\
\cdots \\
\Psi(\text{MA}(f^l_T, F^l, F^l), f^l_T)^{\top}
\end{pmatrix}^{\top}\\
& \Psi(\alpha, \beta) = \text{LayerNorm}(\alpha + \beta) \\
& \text{PF}(\gamma) = M^l_2\max(0, M^l_1 \gamma + b^l_1) + b^l_2
\end{align}
where $\Psi(\cdot)$ represents the function that performs layer normalization on the residual output, $\text{PF}(\cdot)$ denotes the 2-layered feed-forward neural network with ReLU nonlinearity for the first layer, $M_1^l$, $M_2^l$ are the weights for the feed-forward layers, and $b^l_1$, $b^l_2$ are the biases. Notice the self-attention used in eq.~\ref{eq:self_attn}. At each time step $t$, $f^l_t$ is given as the query to the attention layer and the output is the weight sum of $f^l_t$, $t=1,2,...,T$, which encodes not only the information regarding the current time step, but also all other time steps. Therefore, each time step of the output from the self-attention is able to encode all context information. In addition, it is easy to see that the length of the path between time steps is only one. In contrast to recurrent models, this makes the gradient update independent with respect to their position in time, and thus makes learning potential dependencies amongst distant frames easier.

\subsection{Proposal Decoder}
\label{sec:proposal}
Our event proposal decoder is based on ProcNets~\cite{zhou2017procnets}, for its state-of-the-art performance on long dense event proposals. 
We adopt the same anchor-offset mechanism as in ProcNets and design a set of $N$ explicit anchors for event proposals. Each anchor-based proposal is represented by an event proposal score $P_e \in [0,1]$ and two offsets:  center $\theta_c$ and length $\theta_l$. The associated anchor has length $l_a$ and center $c_a$. The proposal boundaries ($S_p$, $E_p$) are determined by the anchor locations and offsets:
\begin{align}
\begin{split}
c_p=c_a+\theta_c l_a \quad & l_p=l_a \exp \{\theta_l\}, \\
S_p=c_p-l_p/2 \quad & E_p=c_p+l_p/2.
\end{split}
\end{align}
These proposal outputs are obtained from temporal convolution (\ie 1-D convolutions) applied on the last layer output of the visual encoder. 
The score indicates the likelihood for a proposal to be an event. The offsets are used to adjust the proposed segment boundaries from the associated anchor locations. We made following changes to ProcNets:

\begin{itemize}
\topsep=0em
\itemsep=0em
\parskip=0em
\parsep=0em
\item The sequential prediction module in ProcNets is removed, as the event segments in a video are not closely coupled and the number of events is small in general.
\item Use input from a multi-head self-attention layer instead of a bidirectional LSTM (Bi-LSTM) layer ~\cite{graves2005framewise}. 
\item Use multi-layer temporal convolutions to generate the proposal score and offsets. The temporal convolutional network contain three 1-D conv. layers, with batch normalization~\cite{ioffe2015batch}. We use ReLU activation for hidden layers.
\item In our model, the conv. stride depends on kernel size ($\ceil{\frac{kernel\;size}{s}}$) versus always 1 in ProcNets\footnote{$s$ is a scalar that affects the convolution stride for different kernel size}.
\end{itemize}

We encode the video context by a self-attention layer as it has potential to learn better context representation. Changing stride size based on kernel size reduces the number of longer proposals so that the training samples is more balanced, because a larger kernel size makes it easier to get good overlap with ground truth. It also speeds up training as the number of long proposals is reduced.

\subsection{Captioning Decoder} 
\label{sec:caption}

\noindent\textbf{Masked Transformer.}\quad The captioning decoder takes input from both the visual encoder and the proposal decoder. Given a proposal tuple $(P_e, S_p, E_p)$ and visual representations $\{F^1,\ldots,F^L\}$, the $L$-layered captioning decoder generates the $t$-th word by doing the following
\begin{align}
&Y^{l+1}_{\le t} = \text{C}(Y^l_{\le t}) = \Psi(\text{PF}(\Phi(Y^l_{\le t})),\Phi(Y^l_{\le t}))\\
&\Phi(Y^l_{\le t}) = 
\begin{pmatrix}
\Psi(\text{MA}(\Omega(Y^l_{\le t})_1, \hat{F}^l, \hat{F}^l), \Omega(Y^l_{\le t})_1)\\
\cdots \\
\Psi(\text{MA}(\Omega(Y^l_{\le t})_{t}, \hat{F}^l, \hat{F}^l), \Omega(Y^l_{\le t})_{t})
\end{pmatrix}\\
&\Omega(Y^l_{\le t}) = 
\begin{pmatrix}
\Psi(\text{MA}(y^l_1, Y^l, Y^l), y^l_1)^{\top}\\
\cdots \\
\Psi(\text{MA}(y^l_t, Y^l, Y^l), y^l_{t})^{\top}
\end{pmatrix}\\
\label{eq:mask}
& \hat{F}^l = f_M(S_p, E_p) \odot F^l\\
\label{eq:word}
& p(w_{t+1}|X, Y^L_{\le t}) = \text{softmax}(W^V y^L_{t+1})
\end{align}
where $y^0_i$ represents word vector, $Y^l_{\le t} = \{y^l_1,\ldots, y^l_t\}$, $w_{t+1}$ denotes the probability of each word in the vocabulary for time $t+1$, $W^V \in \mathbb{R}^{\nu \times d}$ denotes the word embedding matrix with vocabulary size $\nu$, and $\odot$ indicates elementwise multiplication. $\text{C}(\cdot)$ denotes the decoder representation, \ie the output from feed-forward layer in Fig.~\ref{fig:fig1}. $\Phi(\cdot)$ denotes the cross module attention that use the current decoder states to attend to encoder states (\ie multi-head attention in Fig.~\ref{fig:fig1}). $\Omega(\cdot)$ represents the self-attention in decoder. Notice that the subscript ${\le t}$ restricts the attention only on the already generated words.  $f_M : \mathbb{R}^2 \mapsto [0,1]^T$ is a masking function that output values (near) zero when outside the predicted starting and ending locations, and (near) one otherwise. With this function, the receptive region of the model is restricted to the current segment so that the visual representation focuses on describing the current event. Note that during decoding, the encoder performs the forward propagation again so that the representation of each encoder layer contains only the information for the current proposal (see eq. \ref{eq:mask}). This is different from simply multiplying the mask with the existing representation from the encoder during proposal prediction, since the representation of the latter still contains information that is outside the proposal region. The representation from the $L$-th layer of captioning decoder is then used for predicting the next word for the current proposal using a linear layer with softmax activation (see eq.~\ref{eq:word}).

\noindent \textbf{Differentiable Proposal Mask.}\quad
We cannot choose any arbitrary function for $f_M$ as a discrete one would prevent us from doing end-to-end training. We therefore propose to use a fully differentiable function to obtain the mask for visual events.
This function $f_M$ maps the predicted proposal location to a differentiable mask $M\in \mathbb{R}^T$ for each time step $i \in \{1, \ldots, T\}$.
\begin{align}
& f_M(S_p, E_p, S_a, E_a, i) = \sigma(g( \\
& [\rho(S_p, :), \rho(E_p, :), \rho(S_a, :), \rho(E_e, :), \text{Bin}(S_a, E_a, :)]))
\nonumber
\end{align}
\begin{equation}
\rho(pos, i) = \begin{cases}
                     \sin (pos/10000^{i/d}) \;\;\;\qquad i \textit{ is even} \\
                     \cos (pos/10000^{(i-1)/d})\;\;\textit{   otherwise}
                    \end{cases}
\end{equation}
\begin{equation}
\text{Bin}(S_a, E_a, i) = 
     \begin{cases}
     1 \qquad \textit{if } i \in [S_a, E_a]\\
     0 \qquad otherwise
     \end{cases}
\end{equation}
where $S_a$ and $E_a$ are the start and end position of anchor, $[\cdot]$ denotes concatenation, $g(\cdot)$ is a continuous function, and $\sigma(\cdot)$ is the logistic sigmoid function. We choose to use a multilayer perceptron to parameterize $g$. In other words, we have a feed-forward neural network that takes the positional encoding from the anchor and predicted boundary positions and the corresponding binary mask to predict the continuous mask. We use the same positional encoding strategy as in \cite{vaswani2017attention}.

Directly learning the mask would be difficult and unnecessary, since we would already have a reasonable boundary prediction from the proposal module. Therefore, we use a gated formulation that lets the model choose between the learned continuous mask and the discrete mask obtained from the proposal module. More precisely, the gated masking function $f_{GM}$ is
\begin{align}
\nonumber
& f_{GM}(S_p, E_p, S_a, E_a, i) = \\
& P_e \text{Bin}(S_p, E_p, i) + (1-P_e) f_M(S_p, E_p, S_a, E_a, i)
\end{align}
Since the proposal score $P_e \in [0,1]$, it now acts as a gating mechanism. This can also be viewed as a modulation between the continuous and proposal masks, the continuous mask is used as a supplement for the proposal mask in case the confidence is low from the proposal module.

\subsection{Model Learning}
\label{sec:learning}
Our model is fully differentiable and can be trained consistently from end-to-end
The event proposal anchors are sampled as follows. Anchors that have overlap greater than 70\% with any ground-truth segments are regarded as positive samples and ones that have less than 30\% overlap with all ground-truth segments are negative. The proposal boundaries for positive samples are regressed to the ground-truth boundaries (offsets). We randomly sample $U=10$ anchors from positive and negative anchor pools that correspond to one ground-truth segment for each mini-batch.

The loss for training our model has four parts: the regression loss $\mathcal{L}_r$ for event boundary prediction, the binary cross entropy mask prediction loss $\mathcal{L}_m$,  the event classification loss $\mathcal{L}_e$ (\ie prediction $P_e$), and the captioning model loss $\mathcal{L}_c$. The final loss $\mathcal{L}$ is a combination of these four losses,
\begin{align*}
\mathcal{L}_r & = \text{Smooth}_{\ell 1}(\hat{\theta}_c, \theta_c) + \text{Smooth}_{\ell 1}(\hat{\theta}_l, \theta_l)\\
\mathcal{L}_m^i & = \text{BCE}(Bin(S_p, E_p, i),f_M(S_p, E_p, S_a, E_a, i)) \\
\mathcal{L}_e & = \text{BCE}(\hat{P_e}, P_e) \\
\mathcal{L}_c^t & = \text{CE}(\hat{w}_t, p(w_t|X,Y^L_{\le t-1}))\\
\mathcal{L} & = \lambda_1 \mathcal{L}_r + \lambda_2 \sum_i\mathcal{L}_m^i + \lambda_3 \mathcal{L}_e + \lambda_4 \sum_t\mathcal{L}_c^t
\end{align*}
where $\text{Smooth}_{\ell 1}$ is the smooth $\ell_1$ loss defined in~\cite{girshick2015fast}, BCE denotes binary cross entropy, CE represents cross entropy loss, $\hat{\theta}_c$ and $\hat{\theta}_l$ represent the ground-truth center and length offset with respect to the current anchor, $\hat{P}_e$ is the ground-truth label for the proposed event, $\hat{w}_t$ denotes the ground-truth word at time step $t$, and $\lambda_{1\ldots 4} \in \mathbb{R}^+$ are the coefficients that balance the contribution from each loss. 

\noindent\textbf{Simple Single Stage Models.}\quad
The key for our proposed model to work is not the single stage learning of a compositional loss, but the ability to keep the consistency between the proposal and captioning. For example, we could make a single-stage trainable model by simply sticking them together with multi-task learning. More precisely, we can have the same model but choose a non-differentiable masking function $f_M$ in eq.~\ref{eq:mask}. The same training procedure can be applied for this model (see the following section).
Since the masking function would then be non-differentiable, error from the captioning model cannot be back propagated to modify the proposal predictions. However, the captioning decoder is still able to influence the visual representation that is learned from the visual encoder. This may be undesirable, as the updates the visual representation may lead to worse performance for the proposal decoder. As a baseline, we also test this single-stage model in our experiments.

\section{Implementation Details}
For the proposal decoder, the temporal convolutional networks take the last encoding output from video encoder as the input. The sizes of the temporal convolution kernels vary from 1 to 251 and we set the stride factor $s$ to 50. For our Transformer model, we set the model dimension $d=1024$ (same as the Bi-LSTM hidden size) and set the hidden size of feed-forward layer to 2048. We set number of heads (H) to 8. In addition to the residual dropout and attention dropout layers in Transformer, we add a 1-D dropout layer at the visual input embedding to avoid overfitting. We use recurrent dropout proposed in~\cite{gal2016theoretically} for this 1-D dropout. Due to space limits, more details are included in the supplementary material.

\section{Experiments}
\label{sec:exp}
\subsection{Datasets}
ActivityNet Captions~\cite{krishna2017dense} and YouCookII~\cite{zhou2017procnets} are the two largest datasets with temporal event segments annotated and described by natural language sentences. ActivityNet Captions contains 20k videos, and on average each video has 3.65 events annotated. 
YouCookII has 2k videos and the average number of segments per video is 7.70. The train/val/test splits for ActivityNet Captions are 0.5:0.25:0.25 while for YouCookII are 0.66:0.23:0.1. We report our results from both datasets on the validation sets. For ActivityNet Captions, we also show the testing results on the evaluation server while the testing set for YouCookII is not available.

\noindent\textbf{Data Preprocessing.}\quad We down-sample the video every 0.5s and extract the 1-D appearance and optical flow features per frame, as suggested by Xiong et al.~\cite{xiong2016cuhk}. For appearance features, we take the output of the ``Flatten-673'' layer in ResNet-200~\cite{he2016deep}; for optical flow features, we extract the optical flow from 5 contiguous frames, encode with BN-Inception~\cite{ioffe2015batch} and take output of the ``global-pool'' layer. Both networks are pre-trained on the ActivityNet dataset~\cite{caba2015activitynet} for the action recognition task. We then concatenate the two feature vector and further encode with a linear layer.
We set the window size $T$ to 480. The input is zero padded in case the number of sampled frames is smaller than the size of the window. Otherwise, the video is truncated to fit the window.
%
Note that we do not fine-tune the visual features for efficiency considerations, however, allowing fine-tuning may lead to better performance.

\subsection{Baseline and Metrics}
\noindent\textbf{Baselines.}\quad Most of the existing methods can only caption an entire video or specified video clip.
For example, LSTM-YT~\cite{venugopalan2014translating}, S2YT~\cite{venugopalan2015sequence}, TempoAttn~\cite{yao2015describing}, H-RNN~\cite{yu2015video} and DEM~\cite{krishna2017dense}.
The most relevant baseline is TempoAttn, where the model temporally attends on visual sequence inputs as the input of LSTM language encoder. For a fair comparison, we made the following changes to the original TempoAttn.
First, all the methods take the same visual feature input. Second, we add a Bi-LSTM context encoder to TempoAttn while our method use self-attention context encoder. Third, we apply temporal attention on Bi-LSTM output for all the language decoder layers in TempoAttn since our decoder has attention each layer. We name this baseline Bi-LSTM+TempoAttn. Since zero inputs deteriorates Bi-LSTM encoding, we only apply the masking on the output of the LSTM encoder when it is passed to the decoder. 
We also compare with a 
a simple single-stage Masked Transformer baseline as mentioned in section \ref{sec:learning}, where the model employs a discrete binary mask.  

For event proposals, we compare our self-attention transformer-based model with ProcNets and our own baseline with Bi-LSTM. For captioning-only models, we use the same baseline as the full dense video captioning but instead, replace the learned proposals with ground-truth proposals.
%
Results for other dense captioning methods (\eg the best published method DEM~\cite{krishna2017dense}) are not available on the validation set nor is the source code released. So, we compare our methods against those methods that participated in CVPR 2017 ActivityNet Video Dense-captioning Challenge~\cite{ghanem2017activitynet} for test set performance on ActivityNet.

\noindent\textbf{Evaluation Metrics.}\quad For ground-truth segment captioning, we measure the captioning performance with most commonly-used evaluation metrics: BLEU\@\{3,4\} and METEOR. For dense captioning, the evaluate metric takes both proposal accuracy and captioning accuracy into account. Given a tIoU threshold, if the proposal has an overlapping larger than the threshold with any ground-truth segments, the metric score is computed for the generated sentence and the corresponding ground-truth sentence. Otherwise, the metric score is set to 0. The scores are then averaged across all the proposals and finally averaged across all the tIoU thresholds--0.3, 0.5, 0.7, 0.9 in this case.

\subsection{Comparison with State-of-the-Art Methods}
We compare our proposed method with baselines on the ActivityNet Caption dataset. The validation and testing set results are shown in Tab. \ref{tbl:result_densecap} and \ref{tbl:result_leaderboard}, respectively. All our models outperform the LSTM-based models by a large margin, which may be attributed to their better ability of modeling long-range dependencies. 

We also test the performance of our model on the YouCookII dataset, and the result is shown in Tab. \ref{tbl:result_densecap_yc2}. Here, we see similar trend on performance. Our transformer based model outperforms the LSTM baseline by a significant amount. 
However, the results on learned proposals are much worse as compared to the ActivityNet dataset. 
This is possibly because of small objects, such as utensils and ingredients, are hard to detect using global visual features but are crucial for describing a recipe. Hence, one future extension for our work is to incorporate object detectors/trackers~\cite{yu2015compositional,yu2013grounded} into the current captioning system.

\begin{table}[t]
\centering
\caption{Captioning results from ActivityNet Caption Dataset with learned event proposals. All results are on the validation set and all our models are based on 2-layer Transformer. We report BLEU (B) and METEOR (M). All results are on the validation set. Top scores are highlighted.}
\label{tbl:result_densecap}
    {\small
\begin{tabular}{lrrrr}
\toprule
Method & B@3 & B@4 & M \\ 
\midrule
Bi-LSTM & \multirow{2}{*}{2.43} & \multirow{2}{*}{1.01} & \multirow{2}{*}{7.49} \\
+TempoAttn \\
\midrule
Masked Transformer & 4.47 & 2.14 & 9.43 \\
End-to-end Masked Transformer & \textbf{4.76} & \textbf{2.23} & \textbf{9.56} \\
\bottomrule
\end{tabular}
    }
    \vspace{-8pt}
\end{table}

\begin{table}[t]
\centering
\caption{Dense video captioning challenge leader board results. For results from the same team, we keep the highest one.}
\label{tbl:result_leaderboard}
    {\small
\begin{tabular}{lr}
\toprule
Method & METEOR \\
\midrule
DEM~\cite{krishna2017dense} & 4.82 \\
Wang et al. & 9.12 \\
Jin et al. & 9.62 \\
Guo et al. & 9.87 \\
Yao et al.\footnotemark (Ensemble) & 12.84 \\
\midrule
Our Method & \textbf{10.12} \\
\bottomrule
\end{tabular}
    }
    \vspace{-8pt}
\end{table}
\footnotetext{This work is unpublished. It employs external data for model training and the final prediction is obtained from an ensemble of models.}

\begin{table}[t]
\centering
\caption{Recipe generation benchmark on YouCookII validation set. GT proposals indicate the ground-truth segments are given during inference.}
\label{tbl:result_densecap_yc2}
    {\small
\begin{tabular}{lcccc}
\toprule
\multirow{2}{*}{Method} & \multicolumn{2}{c}{\textbf{GT Proposals}} & \multicolumn{2}{c}{\textbf{Learned Proposals}}\\
& B@4 & M & B@4 & M\\
\midrule
Bi-LSTM  & \multirow{2}{*}{0.87} & \multirow{2}{*}{8.15} & \multirow{2}{*}{0.08} & \multirow{2}{*}{4.62} \\ 
+TempoAttn &  &  &  \\  
\midrule
Our Method & \textbf{1.42} & \textbf{11.20} & \textbf{0.30} & \textbf{6.58} \\ 
\bottomrule
\end{tabular}
    }
    \vspace{-8pt}
\end{table}

\begin{figure*}[t]
\centering 
  \includegraphics[width=0.9\textwidth]{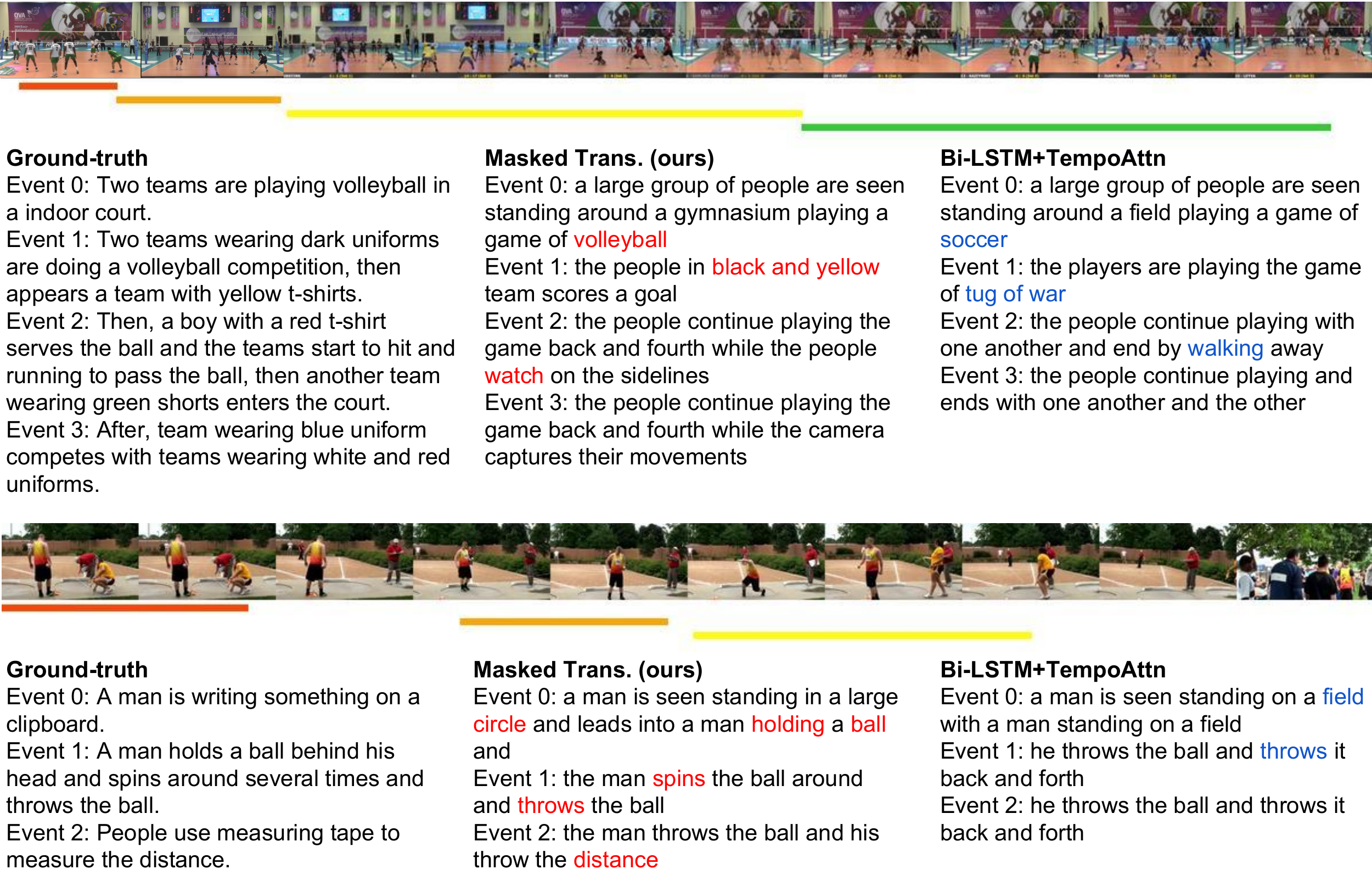}
    \vspace{-2pt}
       \caption{Qualitative results on ActivityNet Captions. The color bars represent different events. Colored text highlight relevant content to the event. Our model generates more relevant attributes as compared to the baseline.}
    \vspace{-8pt}
\label{fig:anet_qual}
\end{figure*}

We show qualitative results in Fig.~\ref{fig:anet_qual} where the proposed method generates captions with more relevant semantic information. More visualizations are in the supplementary.

\subsection{Model Analysis}
In this section we perform experiments to analyze the effectiveness of our model on different sub-tasks of dense video captioning.

\begin{figure}[t]
\centering 
  \includegraphics[width=0.8\columnwidth]{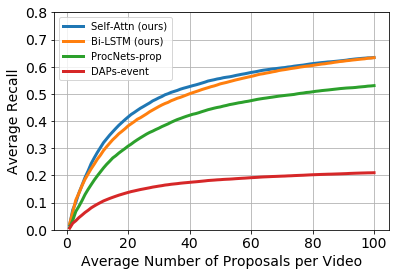}
    \vspace{-2pt}
       \caption{Event proposal recall curve under tIoU threshold 0.8 with average 100 proposals per video.}
    \vspace{-8pt}
\label{fig:prop_recall_curve}
\end{figure}

\noindent\textbf{Video Event Proposal.}\quad We first evaluate the effect of self-attention on event proposal, and the results are shown in Tab.~\ref{tbl:proposal}. We use standard average recall (AR) metric ~\cite{escorcia2016daps,ghanem2017activitynet} given 100 proposals. Bi-LSTM indicates our improved ProcNets-prop model by using temporal convolutional and large kernel strides. We use our full model here, where the context encoder is replaced by our video encoder. We have noticed that the anchor sizes have a large impact on the results. So, for fair comparison, we maintain the same anchor sizes across all three methods. Our proposed Bi-LSTM model gains a 7\% relative improvement from the baseline results from the deeper proposal network and more balanced anchor candidates. Our video encoder further yields a 4.5\% improvement from our recurrent nets-based model. We show the recall curve under high tIoU threshold (0.8) in Fig.~\ref{fig:prop_recall_curve} follow the convention~\cite{krishna2017dense}. DAPs~\cite{escorcia2016daps}, is initially proposed for short action proposals and adapted later for long event proposal~\cite{krishna2017dense}. The proposed models outperforms DAPs-event and ProcNets-prop by significant margins. Transformer based and Bi-LSTM based models yield similar recall results given sufficient number of proposals (100), while our self-attention encoding model is more accurate when the allowed number of proposals is small.

\begin{table}[t]
\centering
\caption{Event proposal results from ActivityNet Captions dataset. We compare our proposed methods with our baseline method ProcNets-prop on the validation set.}
\label{tbl:proposal}
    {\small
\begin{tabular}{lr}
\toprule
Method & Average Recall (\%) \\
\midrule
ProcNets-prop~\cite{zhou2017procnets} & 47.01 \\
Bi-LSTM (ours) & 50.65 \\
Self-Attn (our) & \textbf{52.95} \\
\bottomrule
\end{tabular}
    }
    \vspace{-8pt}
\end{table}

\noindent\textbf{Dense Video Captioning.}\quad Next, we look at the dense video captioning results in an ideal setting: doing the captioning based on the ground-truth event segments. This will give us an ideal captioning performance since all event proposals are accurate. Because we need access to ground-truth event proposal during test time, we report the results on validation set\footnote{The results are overly optimistic, however, it is fine here since we are interested in the best situation performance. The comparison is also fair, since all methods are tuned to optimize the validation set performance.} (see Tab.~\ref{tbl:result_gt_caption}).
The proposed Masked Transformer (section \ref{sec:caption}) outperforms the baseline by a large margin (by more than 1 METEOR point). This directly substantiates the effectiveness of the transformer on both visual and language encoding and multi-head temporal attention. We notice that as the number of encoder and decoder layers increases, the performance gets further boosts by 1.3\%-1.7\%. As can be noted here, the 2-layer transformer strikes a good balance point between performance and computation, and thus we use 2-layer transformer for all our experiments.

\begin{table}[t]
\centering
\caption{Captioning results from ActivityNet Caption Dataset with ground-truth proposals. All results are on the validation set. Top two scores are highlighted.}
\label{tbl:result_gt_caption}
    {\small
\begin{tabular}{lrrr}
\toprule
Method & B@3 & B@4 & M \\ 
\midrule


Bi-LSTM & \multirow{2}{*}{4.8} & \multirow{2}{*}{2.1} & \multirow{2}{*}{10.02} \\
+TempoAttn \\ 
\midrule
\textbf{Our Method} \\
1-layer & \textbf{5.80} & 2.66 & 10.92 \\ 
2-layer & 5.69 & 2.67 & 11.06 \\ 
4-layer & \textbf{5.70} & \textbf{2.77} & \textbf{11.11} \\ 
6-layer & 5.66 & \textbf{2.71} & \textbf{11.10} \\ 
\bottomrule
\end{tabular}
    }
    \vspace{-10pt}
\end{table}

\begin{table}[t]
\centering
\caption{Evaluating only long events from ActivityNet Caption Dataset. GT proposals indicate the ground-truth segments are given during inference.}
\label{tbl:long_events_only}
    {\small
\begin{tabular}{lcccc}
\toprule
& \multicolumn{2}{c}{\textbf{GT Proposals}} & \multicolumn{2}{c}{\textbf{Learned Proposals}} \\
Method & B@4 & M & B@4 & M \\
\midrule
Bi-LSTM & \multirow{2}{*}{0.84} & \multirow{2}{*}{5.39} & \multirow{2}{*}{0.42} & \multirow{2}{*}{3.99} 
\\
+TempoAttn \\
\midrule
Our Method & \textbf{1.13} & \textbf{5.90} & \textbf{1.04} & \textbf{5.93} \\ 
\bottomrule
\end{tabular}
    }
    \vspace{-10pt}
\end{table}

\noindent\textbf{Analysis on Long Events.}\quad As mentioned in section \ref{sec:encoder}, learning long-range dependencies should be easier with self-attention, since the next layer observes information from all time steps of the previous layer. To validate this hypothesis directly, we test our model against the LSTM baseline on longer event segments (where the events are at least 50s long) from the ActivityNet Caption dataset, where learning the long-range dependencies are crucial for achieving good performance. It is clear from the result (see Tab. \ref{tbl:long_events_only}) that our transformer based model performs significantly better than the LSTM baseline. The discrepancy is even larger when the model needs to learn both the proposal and captioning, which demonstrate the effectiveness of self-attention in facilitate learning long range dependencies.

\section{Conclusion}
We propose an end-to-end model for dense video captioning. The model is composed of an encoder and two decoders. The encoder encodes the input video to proper visual representations. The proposal decoder then decodes from this representation with different anchors to form video event proposals. The captioning decoder employs a differentiable masking network to restrict its attention to the proposal event, ensures the consistency between the proposal and captioning during training. In addition, we propose to use self-attention for dense video captioning. We achieved significant performance improvement on both event proposal and captioning tasks as compared to RNN-based models. We demonstrate the effectiveness of our models on ActivityNet Captions and YouCookII dataset.

\small{\noindent\textbf{Acknowledgement.} The technical work was performed while Luowei was an intern at Salesforce Research. This work is also partly supported by ARO W911NF-15-1-0354 and DARPA FA8750-17-2-0112. This article solely reflects the opinions and conclusions of its authors but not the funding agents.}

{\small
\bibliographystyle{ieee}
\bibliography{densecap}
}

\clearpage
\section{Appendix}

\subsection{Implementation Details}
The sizes of the temporal convolution kernels in the proposal module are 1 2, 3, 4, 5, 7, 9, 11, 15, 21, 29, 41, 57, 71, 111, 161, 211 and 251. We set the hyper-parameters for End-to-end Masked Transformer as follows. The dropout ratio for Transformer is set to 0.2 and that for visual input embedding is set to 0.1. We set the loss coefficients $\lambda_1, \lambda_2, \lambda_3, \lambda_4$ to $10, 1, 1, 0.25$. 
For training, we use stochastic gradient descent (SGD) with Nesterov momentum, the learning rate is set between 0.01 and 0.1 depending on the convergence, and the momentum is set at 0.95. We decay the learning rate by half on plateau. We also clip the gradient~\cite{pascanu2013difficulty} to have global $\ell_2$ norm of 1. 
For inference, we first pick event proposals with prediction score higher than a pre-defined threshold (0.7). We remove proposals that have high overlap (\ie $\ge 0.9$) with each other. For each video, we have at least 50, and at most 500 event proposals. 
The descriptions are then generated for each of the proposal, and we use greedy decoding for text generation with at most 20 words.
We implement the model in PyTorch and train it using 8 Tesla K80 GPUs with synchronous SGD. The model typically takes a day to converge.

The implementation for proposal-only and captioning-only model is slightly different. We apply Adam for training rather than SGD and set the learning rate to 0.0001. When training the captioning-only model, we apply scheduled sampling~\cite{bengio2015scheduled}. We set the sampling ratio to 0.05 at the beginning of training, and increase it by 0.05 every 5 epoch until it reaches 0.25. Note that applying scheduled sampling to End-to-end Masked Transformer yield no improvements and hence disabled. In the proposal-only model, we report the results on a single-layer Transformer with the model size and hidden size to be 512 and 128. The temporal conv. stride factor $s$ is set to 10. 

\subsection{Additional Results}
To see the effectiveness of self-attention, we performed additional ablation studies, where we apply self-attention module at the encoder or decoder of the LSTM-based baseline. From the result it is clear that self-attention have significant impact on the performance of the model (see Tab. \ref{tbl:rebuttal_anet}), especially as in the language decoder.

Note that the performance of captioning models over ground-truth segments vary little from number of layers. We choose to use 2-layer transformer for the rest of the experiments because 1) the 4 and 6 layer models are more computational expensive; 2) the learning is more complicated when the learned proposals are approximate, and a 2-layer model give us more flexibility for handling this case (see Tab.~\ref{tbl:rebuttal_anet} for results on the 1-layer model).

Self-attention facilitates the learning of long-range dependencies, which should not hurt the performance on modeling relative short-range dependencies. To validate this we tested our model on shorter activities, where the activities are at most 15 seconds long. The result is shown in Tab. \ref{tbl:short_events_only}.

\subsection{Additional Qualitative Results}
\begin{figure*}
\centering
\begin{minipage}{0.7\linewidth}
  \centerline{\includegraphics[width=1\textwidth]{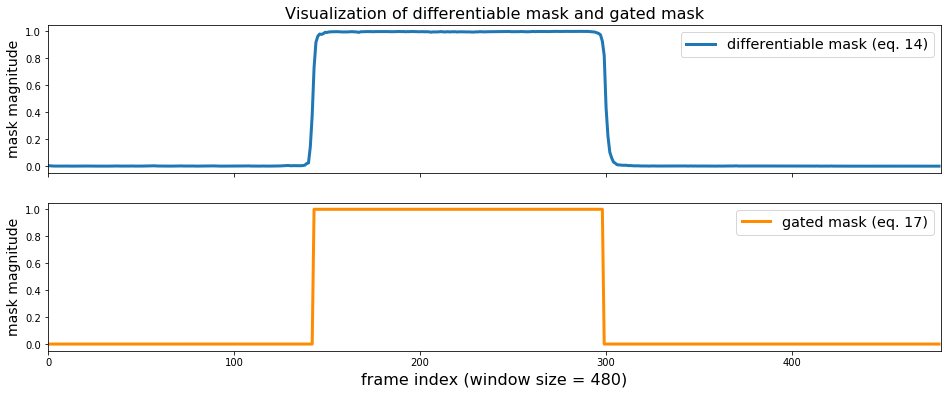}}
  \centerline{\small{(a) High proposal score.}}
\end{minipage}
\vfill
\begin{minipage}{0.7\linewidth}
  \centerline{\includegraphics[width=1\textwidth]{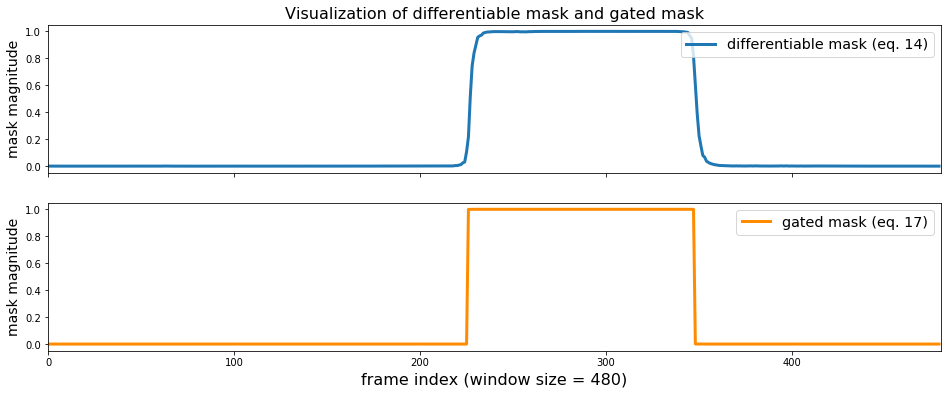}}
  \centerline{\small{(b) High proposal score}}
\end{minipage}
\vfill
\begin{minipage}{0.7\linewidth}
  \centerline{\includegraphics[width=1\textwidth]{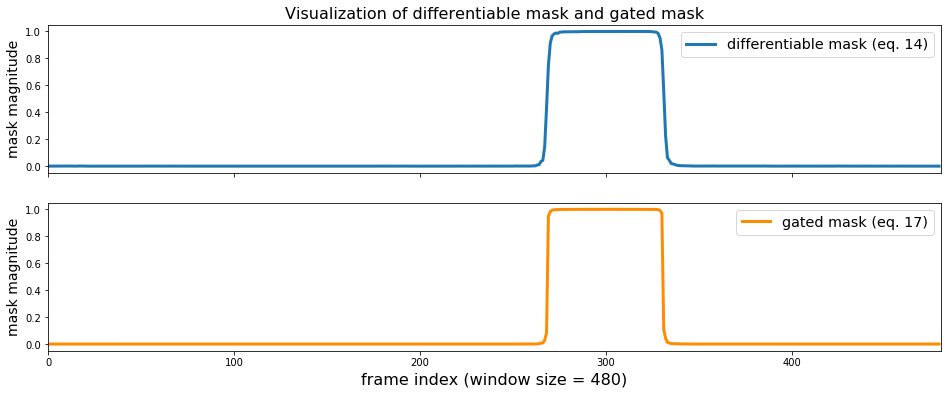}}
  \centerline{\small{(b) Low proposal score}}
\end{minipage}
\vfill
\begin{minipage}{0.7\linewidth}
  \centerline{\includegraphics[width=1\textwidth]{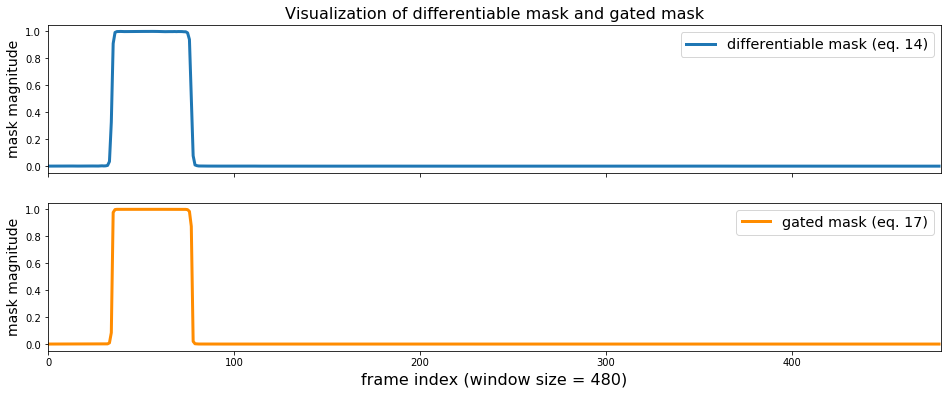}}
  \centerline{\small{(b) Low proposal score}}
\end{minipage}
\caption{Visualization of differentiable masks and final masks under hight (a and b) and low proposal score (c and d). Videos from ActivityNet Captions validation set.} \label{fig:vis_mask}
\end{figure*}

\begin{figure*}[t]
\centering 
  \includegraphics[width=1.0\textwidth]{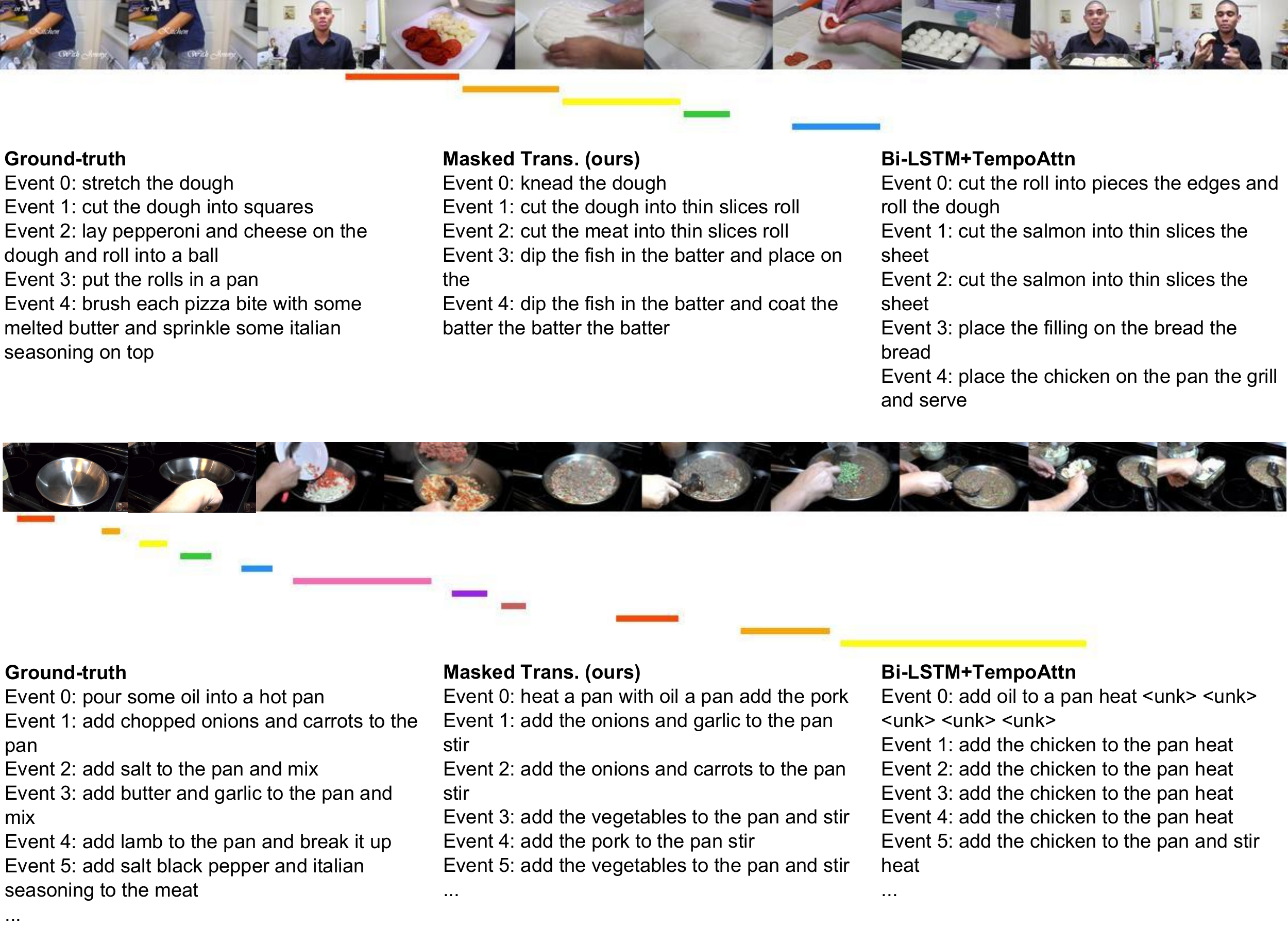}
       \caption{Qualitative results on YouCookII videos. We only showed result for the first 6 events in the second example.}
\label{fig:yc2_qual}
\end{figure*}

\begin{figure}[t]
\centering 
  \includegraphics[width=\columnwidth]{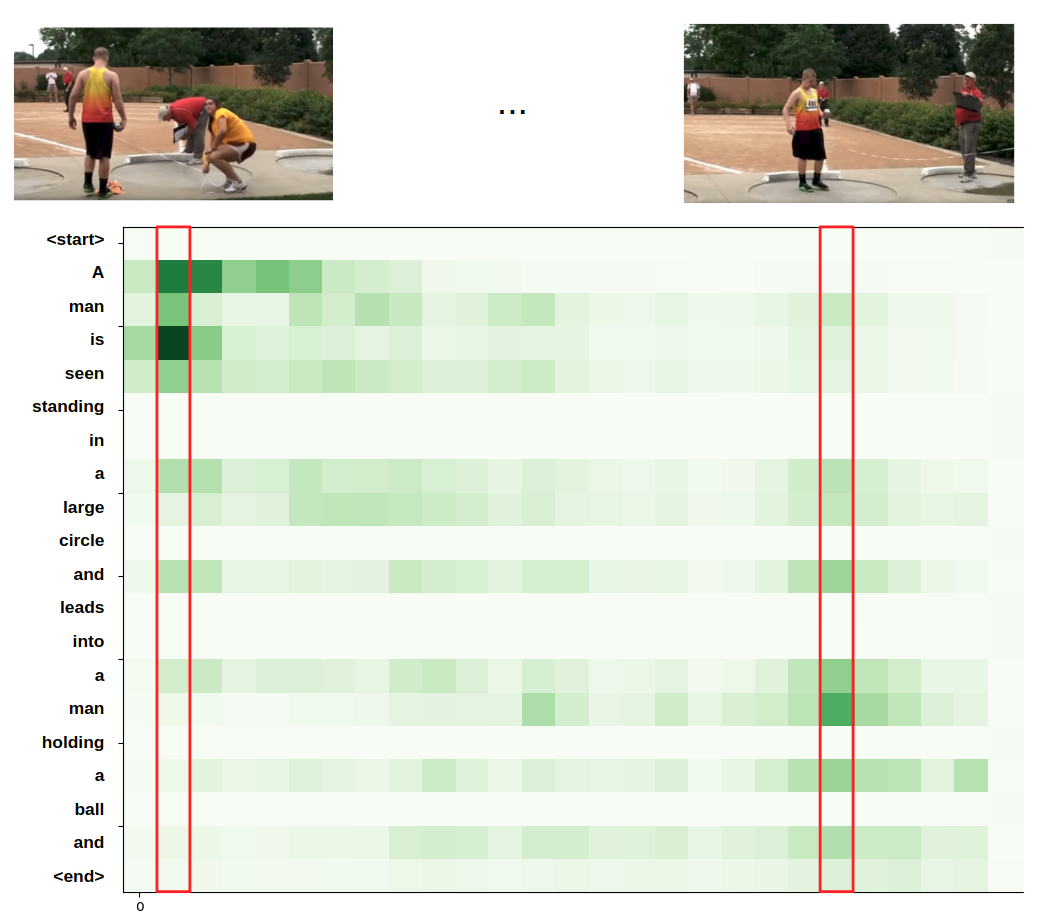}
       \caption{Visualization of weights from the cross-module attention layer. X axis represents the generated words at each time step. Y axis indicates the sampled frames.}
\label{fig:vis_mm_attn}
\end{figure}

We visualize the learned masks in Fig.~\ref{fig:vis_mask}. The first two correspond to the case where the proposal prediction is confident, i.e., proposal scores are high ($>0.9$) and the last two correspond to the case where the prediction is less confident, i.e., proposal scores are low ($<0.7$).
We visualize the cross module attention in Fig.~\ref{fig:vis_mm_attn}. For convenience, we randomly choose one of the attention matrices from the multi-head attention. Also, we notice that attention weights from higher-level self-attention layer is tend to be flatter than these from the lower-level layer.
Qualitative results for YouCookII are shown in Fig.~\ref{fig:yc2_qual}. The visual recognition is challenging result from the small and ambiguous objects (e.g., black pepper, lamb).

\begin{table}[t]
\centering
\caption{Additional ablation experiments on ActivityNet.}
\label{tbl:rebuttal_anet}
    {\small
\begin{tabular}{lrrr}
\toprule
Method & B@3 & B@4 & M \\
\midrule
SelfAttn + LSTM TempoAttn & 2.91 & 1.35 & 7.88 \\ 
BiLSTM + SelfAttn & 4.06 & 1.92 & 9.05 \\
Our Method (1-layer)  & 4.49 & 2.10 & 9.27 \\ 
\bottomrule
\end{tabular}
    }
\end{table}

\begin{table}[t]
\centering
\caption{Evaluating only short events from ActivityNet.}
\label{tbl:short_events_only}
    {\small
\begin{tabular}{lcccc}
\toprule
& \multicolumn{2}{c}{\textbf{GT Proposals}} & \multicolumn{2}{c}{\textbf{Learned Proposals}} \\
Method & B@4 & M & B@4 & M \\
\midrule
Bi-LSTM+TempoAttn & 0.74 & 5.29 & 0.23 & 4.43 \\
\midrule
Our Method & 0.87 & 5.82 & 0.68 & 5.06 \\ 
\bottomrule
\end{tabular}
    }
\end{table}

\end{document}